\documentclass[10pt,twocolumn,letterpaper]{article}

\usepackage{cvpr}
\usepackage{times}
\usepackage{epsfig}
\usepackage{graphicx}
\usepackage{amsmath}
\usepackage{amssymb}

\usepackage{enumitem} 
\usepackage{verbatim} 
\usepackage{xcolor} 
\usepackage{subfigure} 
\usepackage{multirow}
\usepackage{bigstrut}
\usepackage{booktabs}
\usepackage{comment}

\usepackage[pagebackref=true,breaklinks=true,letterpaper=true,colorlinks,bookmarks=false]{hyperref}

\cvprfinalcopy 

\def\cvprPaperID{2040} 

\ifcvprfinal\fi

\begin{document}

\title{Multi-Granularity Reference-Aided Attentive Feature Aggregation\\ for Video-based Person Re-identification}

\author{Zhizheng Zhang{$^1$}\thanks{This work was done when Zhizheng Zhang was an intern at MSRA.} \qquad 
Cuiling Lan$^2$\thanks{Corresponding author.}
\qquad 
Wenjun Zeng$^2$ 
\qquad 
Zhibo Chen$^{1\dagger}$
\and 
\normalsize
	$^1$University of Science and Technology of China \qquad $^2$Microsoft Research Asia\\
	{\tt\small zhizheng@mail.ustc.edu.cn} \quad {\tt\small \{culan,wezeng\}@microsoft.com} \quad {\tt\small chenzhibo@ustc.edu.cn}
}

\maketitle
\thispagestyle{empty}

\begin{abstract}
    Video-based person re-identification (reID) aims at matching the same person across video clips. It is a challenging task due to the existence of redundancy among frames, newly revealed appearance, occlusion, and motion blurs. In this paper, we propose an attentive feature aggregation module, namely \textbf{M}ulti-\textbf{G}ranularity \textbf{R}eference-aided \textbf{A}ttentive \textbf{F}eature \textbf{A}ggregation (MG-RAFA), to delicately aggregate spatio-temporal features into a discriminative video-level feature representation. In order to determine the contribution/importance of a spatial-temporal feature node, we propose to learn the attention from a global view with convolutional operations. Specifically, we stack its relations, \ie, pairwise correlations with respect to a representative set of reference feature nodes (S-RFNs) that represents global video information, together with the feature itself to infer the attention. Moreover, to exploit the semantics of different levels, we propose to learn multi-granularity attentions based on the relations captured at different granularities. Extensive ablation studies demonstrate the effectiveness of our attentive feature aggregation module MG-RAFA. Our framework achieves the state-of-the-art performance on three benchmark datasets.
\end{abstract}

\section{Introduction}

\begin{figure}[t]
	\begin{center}
		\includegraphics[width=1\linewidth]{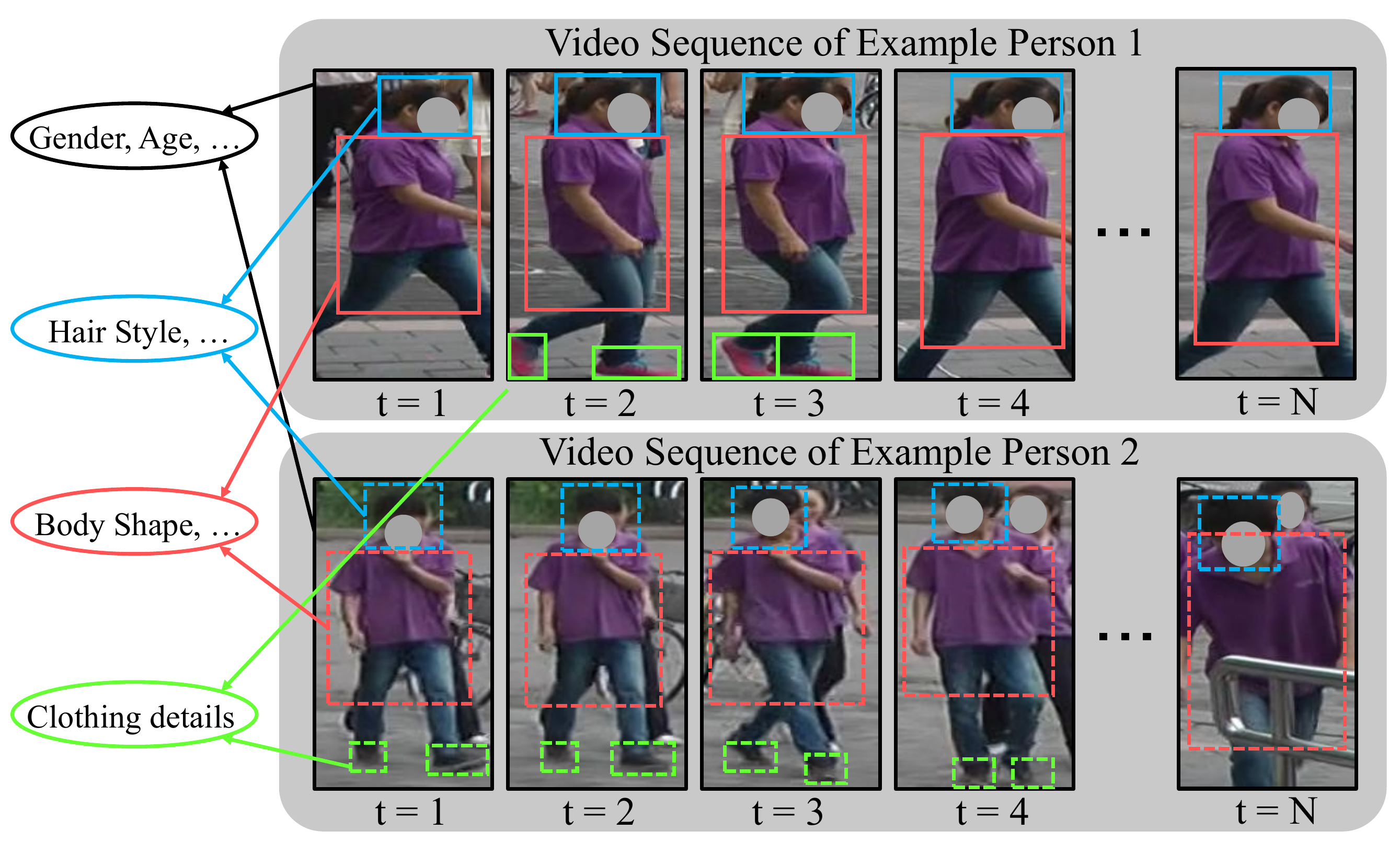}
	\end{center}
	\vspace{-4mm}
	\caption{Illustration of two video sequences \protect\footnotemark[1] of different identities. We observe that: (a) videos have redundancy with repetitive contents spanning over time; (b) there are some contents which occasionally appear but are discriminative factors (such as red shoes of person 1 in $t=2,3$). (c) discriminative factors/properties could be captured at different granularities/scales (\eg, body shape can be captured from a large region (coarse granularity) while hair style is captured from small local region (fine granularity).}
	\label{fig:motivation}
	\vspace{-5mm}
\end{figure}

Person re-identification (reID) aims at matching persons in different positions, times, and camera views. Many researches focus on image-based setting by comparing the still images \cite{zhao2017deeply,liu2017end,kalayeh2018human,wang2018mancs,li2018harmonious,liu2018spatial}. With the prevalence of video capturing systems, person reID based on video offers larger capacity for achieving more robust performance. As illustrated in Figure~\ref{fig:motivation}, for a video clip, the visible contents of different frames differ but there are also overlaps/redundancy. In general, the multiple frames of a video clip/sequence can provide more comprehensive information of a person for identification, but also raise more challenges, such as the handling of the presence of abundant redundancy, occlusion, motion blurs.

\footnotetext[1]{All faces in the images are masked for anonymization.}

A typical video-based person reID pipeline \cite{mclaughlin2016recurrent,liu2017quality,xu2017jointly,gao2018revisiting,li2018diversity,fu2019sta,zhao2019attribute,liu2019spatial} extracts and aggregates spatial and temporal features to obtain a single feature vector as the video representation. To make the video-level feature representation precise, comprehensive and discriminative, we should latch onto the informative features from a global view, and meanwhile, remove the interference.

Attention, which aims at strengthening the important features while suppressing the irrelevant ones, matches the aforementioned goal well. Some works have studied spatio-temporal attention \cite{li2018diversity, fu2019sta} or the attentive recurrent neural networks \cite{zhou2017see, liu2018spatial} to aggregate spatial and temporal features. They learn the attention weights for spatial and temporal dimensions separately or sequentially \cite{zhou2017see,li2018diversity}. However, due to the lack of a global view, they suffer from the difficulty in precisely determining whether a feature of some position is important and what the degree of redundancy is within the entire video clip. A diversity regularization is adopted in \cite{li2018diversity, fu2019sta} to remedy this issue, but only alleviates it to some extent. A powerful model is expected which jointly determines the importance levels of each spatio-temporal feature from a global view. Besides, as shown in Figure \ref{fig:motivation}, discriminative factors/semantics could be captured at different granularities (regions of different sizes) by human. However, there is a lack of effective mechanisms to explore such hierarchical characteristics.

In this paper, we propose a Multi-Granularity Reference-aided Attentive Feature Aggregation (MG-RAFA) scheme for video-based person reID. For effective feature aggregation of the spatial and temporal positions, we determine the importance of each feature position/node from a global view, and consider the hierarchy of semantics during this process. For each feature position, we use its relations/affinities with respect to all reference feature nodes, which represent the global structural information (clustering-like patterns), together with the feature itself (appearance information) to model and infer the attention weights for aggregation. This is in part inspired by Relation-aware Global Attention (RGA) \cite{zhangRGA} designed for effective image feature extraction. However, a 3D video is rather different from a 2D image, where a video clip generally presents abundant redundancy along the time dimension, and the spatio-temporal structure patterns are complicated due to the diversity of human poses. 

Considering the characteristics of video, we propose to construct a small but representative set of reference feature nodes (S-RFNs) for globally modelling the pairwise relations, instead of using all the original feature nodes. S-RFNs provides a simplified but representative reference for modeling global relations, which not only eases the difficulty of attention learning but also reduces the computational complexity. Moreover, we also take into account that the semantics are diverse in their granularities as illustrated in Figure~ \ref{fig:motivation}. We propose to hierarchically model relations for attentive feature aggregation, which allows attention learning to be more precise and adaptive with low computational complexity.

In summary, we have three main contributions: 

\begin{itemize}[noitemsep,nolistsep,leftmargin=*]

\item For video-based person reID, we propose a simple yet effective Multi-Granularity Reference-Aided Attention Feature Aggregation (MG-RAFA) module for the joint spatial and temporal attentive feature aggregation.

\item To better capture the discriminative semantics at different granularities, we exploit the relations at multiple granularities to infer attention for feature aggregation.

\item We propose to build a small but representative reference set for more effective relation modeling by compressing the redundancy information of video data. 

\end{itemize}
We conduct extensive experiments to evaluate our proposed feature aggregation for video-based person reID and demonstrate the effectiveness of each technical component. The final system significantly outperforms the state-of-the-art approaches on three benchmark datasets. Besides, the proposed multi-granularity module MG-RAFA further reduces the computational complexity as compared to its single-granularity version SG-RAFA via our innovative design. Our final scheme only slightly increases the computational complexity over the baseline ($<1\%$).


\section{Related Work}

In many practical scenarios, video is ready for access and contains more comprehensive information than a single image. Video-based person reID offers larger optimization space for achieving high reID performance and attracts more and more interests in recent years.

\noindent\textbf{Video-based Person ReID.} Some works simply formulate the video-based person reID problem as an image-based reID problem, which extracts the feature representation for each frame and aggregates the representations of all the frames using temporal average pooling \cite{suh2018part,gao2018revisiting}. McLaughlin \etal apply Recurrent Neural Network on the frame-wise features extracted from CNN to allow information to flow among different frames, and then temporally pool the output features to obtain the final feature representation \cite{mclaughlin2016recurrent}. Similarly, Yan \etal leverage LSTM network to aggregate the frame-wise features to obtain a sequence-level feature representation \cite{yan2016person}. Liu \etal propose a two-stream network in which motion context together with appearance features are accumulated by recurrent neural network \cite{liu2018video}. Inspired by the exploration of 3D Convolutional Neural Network for learning the spatial-temporal representation in other video-related tasks such as action recognition \cite{ji20123d,carreira2017quo}, 3D convolution networks are used to extract sequence-level feature \cite{liao2018video,li2019multi}. These works treat the features with the same importance even though the features for different spatial and temporal positions have different contribution/importance levels for video-based person reID.  

\noindent\textbf{Attention for Image-based Person ReID.} For image-based person reID, many attention mechanisms have been designed to emphasize important features and suppress irrelevant ones for obtaining discriminative features. Some works use the human part/pose/mask information to infer the attention regions for extracting part/foreground features \cite{song2018mask,kalayeh2018human,xu2018attention,song2018mask}. Some works learn the attention in terms of spatial positions or channels in end-to-end frameworks \cite{liu2017end,zhao2017deeply,li2018harmonious,wang2018mancs,yang2019attention}. In \cite{li2018harmonious}, spatial attention and channel attention are adopted to modulate the features. In general, convolutional layers with limited receptive fields are used to learn the spatial attention. Zhang \etal propose a relation-aware global attention to globally learn attention by exploiting the pairwise relations \cite{zhangRGA} and achieve significant improvement for image-based person reID. Despite the wide exploration in image-based reID, attention designs are under-explored for video-based reID, with much fewer efforts on the globally derived attention.

In this paper, motivated in part by \cite{zhangRGA} which is designed for effective feature learning of an image by exploring relations, we design a multi-granularity reference-aided attentive feature aggregation scheme for video-based person reID. Particularly, to compute the relations, we build a small set of  reference nodes instead of using all the nodes for robustness and computational efficiency. Moreover, multi-granularity attention is designed to capture and explore the semantics of different levels.

\noindent\textbf{Attention for Video-based Person ReID.} For video-based person reID, some attention mechanisms have been designed. One category of works considers the mutual influences between the pair of sequences to be matched \cite{xu2017jointly,si2018dual,chen2018video}. In \cite{xu2017jointly}, the temporal attention weights for one sequence was guided by the information from distance matching with the other sequence. However, given a sequence in the gallery set, it needs to prepare different features corresponding to different query images, which is complicated and less friendly in practical application. 

\begin{figure*}[th]
\vspace{-2mm}
\centering
\subfigure[Our pipeline with reference-aided attentive feature aggregation.] 
{\includegraphics[scale=.38]{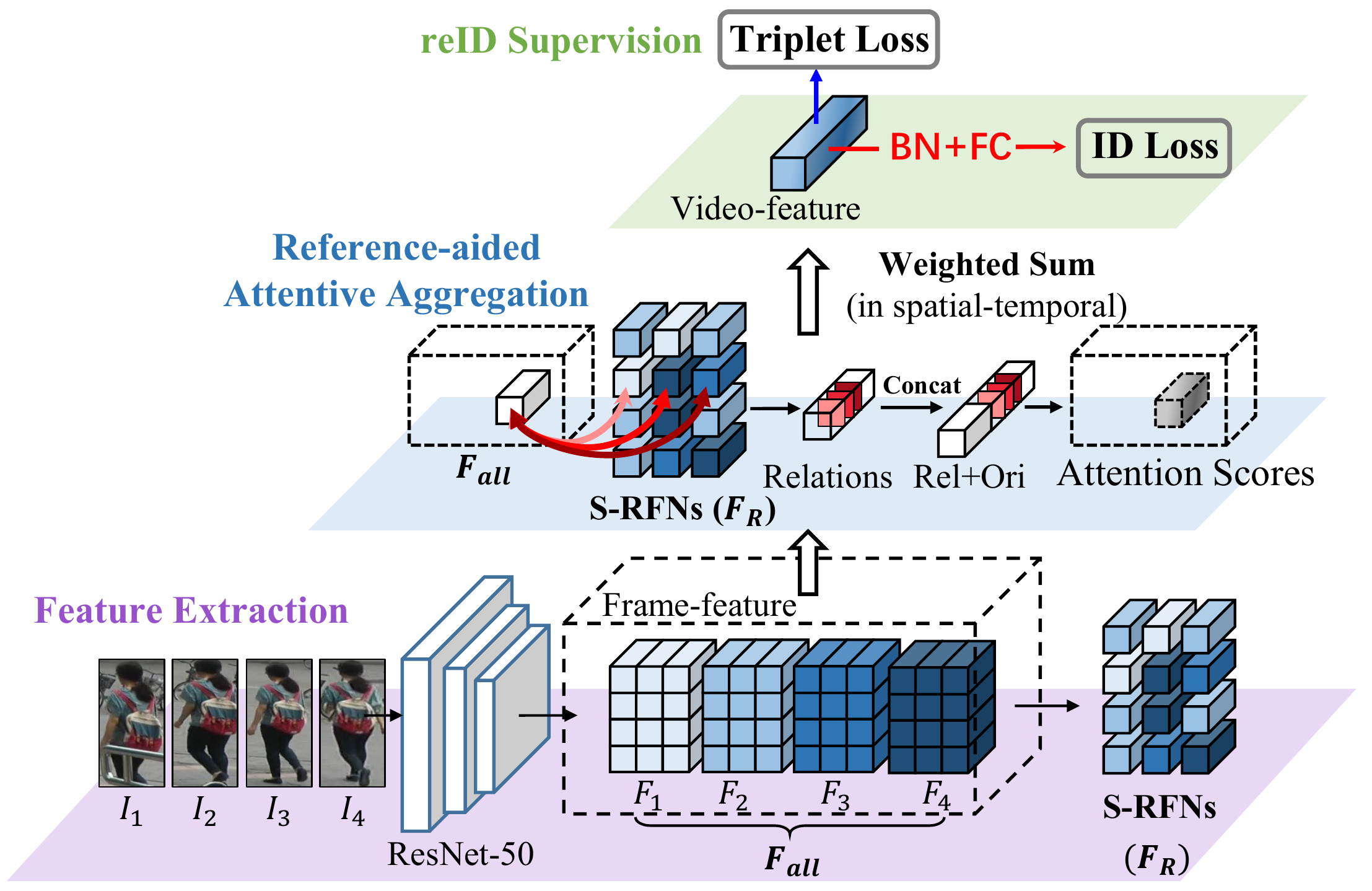}}
\hspace{0.4cm}
\subfigure[Architecture of multi-granularity reference-aided attention.]
{\includegraphics[scale=.38]{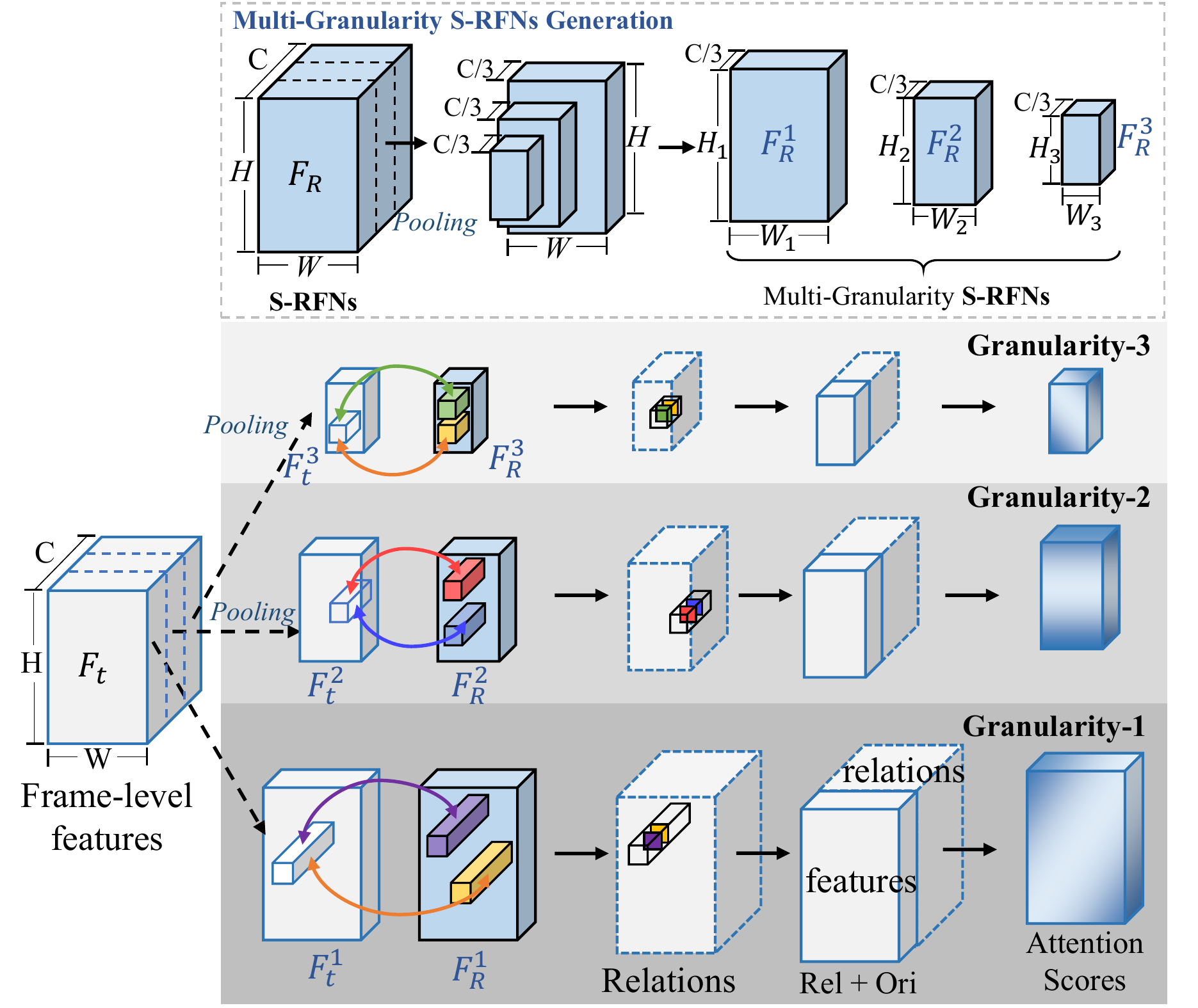}}
\caption{Our proposed Multi-Granularity Reference-aided Attentive Feature Aggregation scheme for video-based person reID. (a) illustrates the reID pipeline with reference-aided attentive feature aggregation. Here, we use four frames ($T=4$) as an example. For clarity, we only show the single-granularity setting in (a) and illustrate the procedure for deriving multiple-granularity reference-aided attention in (b). We use three granularity ($N$=3) here for illustration.}
\label{fig:framework}
\vspace{-3mm}
\end{figure*}

Another category of works independently determines the features of the sequence itself. To weaken the influence of noisy frames, Liu \etal propose a quality aware network (QAN), which estimates the quality score of each frame for aggregating the temporal features as the final feature  \cite{liu2017quality}. Zhou \etal use learned temporal attention weights to update the current frame features as input of RNN \cite{zhou2017see}. Zhao \etal disentangle the features of each frame to semantic attribute related sub-features and re-weight them by the confidence of attribute recognition for temporal aggregation \cite{zhao2019attribute}. These works do not simultaneously generate spatial and temporal attention from a global view for feature aggregation. Recently, spatial-temporal map which is directly calculated from the feature maps is used to aggregate the frame-level feature maps, without using any additional parameters \cite{fu2019sta}. However, as attention is computed in a pre-defined manner, the optimization space is limited.

Considering the lack of effective joint spatio-temporal attention mechanisms for the feature aggregation in video-based person reID, we intend to address this by proposing a multi-granularity reference-aided global attention, which jointly determines the spatial and temporal attention for feature aggregation.

\section{Multi-Granularity Reference-aided Attentive Feature Aggregation}

We propose an effective attention module, namely Multi-Granularity Reference-aided Global Attention (MG-RAFA), for spatial-temporal feature aggregation to obtain a video-level feature vector. In  Section \ref{overview}, we introduce the preliminary. Then, we describe our proposed reference-aided attentive feature aggregation under single-granularity setting in Section \ref{subsec:rga} and elaborate the multi-granularity (our final design of MG-RAFA) in Section \ref{subsec:MG}. We finally present the loss functions in Section \ref{loss}.

\subsection{Overview}
\label{overview}

For video-based person reID, we aim at designing an attentive feature aggregation module that can comprehensively capture discriminative information and exclude interference from a video which in general contains redundancy, newly revealed contents, occlusion and blurring. To achieve this goal, a joint determination of attention for the spatio-temporal features from a global view is important for robust performance.

We propose to learn the attention for each spatio-temporal position/node by exploring the global relations with respect to a set of reference feature nodes. Particularly, for the global relations modeling of a target node, we build a small set of representative feature nodes as reference, instead of using all the feature nodes, to ease optimization difficulty and reduce the computational complexity. Moreover, the discriminative information may physically spread over different semantic levels as illustrated in Figure \ref{fig:motivation}. We thus introduce hierarchical (multi-granularity) relation modeling to capture the semantics at different granularities.

Figure \ref{fig:framework} gives an illustration of our overall framework. For a video tracklet, we sample $T$  frames as $\mathcal{V}=\{I_1, I_2,\cdots, I_T\}$. Through a single frame feature extractor (\eg, ResNet-50 backbone), we obtain a set of feature maps $F_{all}=\{F_t | t=1,2,\cdots,T\}$, where $F_t \in \mathbb{R}^{H \times W \times C}$ includes $H \times W$ feature nodes, ($H$, $W$, $C$ represent the height, width and number of channels, respectively). Based on the proposed multi-granularity reference-aided attention, all feature nodes in this set are weighted summed into a feature vector $\boldsymbol{v}$ as the final video-level feature representation for matching by $l_2$ distance. For clarity, we first present our proposed reference-aided attentive feature aggregation under the single-granularity setting in Subsection \ref{subsec:rga} and introduce the multi-granularity version in Subsection \ref{subsec:MG}.

\subsection{Reference-aided Attentive Feature Aggregation}
\label{subsec:rga}

The extracted feature set $F_{all}=\{F_t | t=1,2,\cdots,T\}$ consists of $K=H \times W \times T$ feature nodes, each of which is a $C$-dimensional feature vector. To determine the importance level of a feature node, it would be helpful if all the other feature nodes are also ``seen", since intuitively people can determine the relative importance of something by comparing it with all others. For a feature node, to determine its importance level, we prepare its relation/affinity with every node as the ingredient to infer the attention. For any node $i$, when stacking its relations with all nodes (\eg, in raster scan order), the number of relation elements is $D=H\times W \times T$.

Taking into account the existence of appearance variations (\eg, caused by pose, viewpoint variations) and large redundancy among frames, the distribution space for relation vectors is large and may cause difficulty in mining the patterns for accurately determining the attention. Thus, we propose to ease the difficulty by choosing a small set of representative feature nodes, instead of all the nodes, as the reference for modeling relations. As we know, for a video tracklet, there is usually large redundancy across temporal frames. For video action recognition, Bobick \etal \cite{bobick2001recognition} propose to use a static vector-image where the vector value at each point is a function of the motion properties at the corresponding spatial location of a video sequence to  compactly represent the information of a video. Motivated by this, we adopt average pooling along the temporal frames to fuse $F_{all}=\{F_t|t=1,\cdots,T\}$ into a feature map $F_R \in \mathbb{R}^{H \times W \times C}$. Different from action recognition where the motion/temporal evolution is important, the temporal motion and evolution in general has no discriminative information for person ReID while the appearances are the key. Thus we simply average the temporal frames to obtain $F_R$ as the reference, \ie, the representative set of reference feature nodes (S-RFNs), to model global relations, which consists of $D=H\times W$ feature nodes. 

For a feature node $\boldsymbol{x}^{i} \in \mathbb{R}^C$ in $F_{all}$, we calculate the relations/affinities between it and all feature nodes in the reference set $F_R$ to model its corresponding relations. A pairwise relation is formulated as the correlation of the two nodes in embedding spaces:
\begin{equation}\label{eq:relation}
r_{i,j} = r(\boldsymbol{x}^{i}, \boldsymbol{y}_{r}^{j})=\mu(\boldsymbol{x}^{i})^{T}\cdot\nu(\boldsymbol{y}_{r}^{j}),
\end{equation}
where $\boldsymbol{y}_{r}^{j} \in \mathbb{R}^C$ denotes a feature node in the reference set $F_R$, $i$ (and $j$) identifies the node index. We define  $\mu(\boldsymbol{x}^{i})=ReLU(W_\mu\boldsymbol{x}^{i})$ and $\nu(\boldsymbol{y}_{r}^{j})=ReLU(W_\nu\boldsymbol{y}_{r}^{j})$, where $W_\mu \in \mathbb{R}^{(C/s)\times C}$ and $W_\nu \in \mathbb{R}^{(C/s)\times C}$ are learned weight matrices, where $s$ is a positive integral which controls the dimension reduction ratio. We implement it by adopting a $1\times 1$ convolutional filter followed by Batch Normalization (BN) and ReLU activation, respectively. Note that we omit BN operations to simplify the notation. By stacking the pairwise relations of the feature node $\boldsymbol{x}^i$ with all the nodes (\eg, scanned in raster scan order) in the reference set $F_R$, we have the relation vector as
\begin{equation}\label{eq:relationvector}
\mathbf{r}_i = [r_{i,1}, r_{i,2},\cdots,r_{i,D}] \in \mathbb{R}^D,
\end{equation}
which compactly reflects the global and clustering-like structural information. In addition, since the relations are stacked into a vector with a fixed scanning order with respect to the reference nodes, the spatial geometric information is also contained in the relation vector.

Intuitively, a person can have a sense of the importance levels of a node once he or she obtains the affinity/correlation of this node with many other ones. Similarly, the relation vector which describes the affinity/relation with all reference nodes provides valuable structural information.    
Particularly, the original feature $\boldsymbol{x}^{i}$ represents local appearance information while the relation feature $\boldsymbol{r}^{i}$ models global relations. They complement and reinforce each other but in different semantic spaces. We thereby combine them together in their respective embedding space and jointly learn, model, and inference the level of importance (attention scores) of the feature node $\boldsymbol{x}^{i}$ through a modeling function as
\begin{equation}\label{eq:attentionscore}
\boldsymbol{a}^{i}=\theta([\phi(\boldsymbol{x}^{i}), \psi(\boldsymbol{r}^{i})]),
\end{equation}
where $\phi(\cdot)$ and $\psi(\cdot)$ are two embedding functions, $[\cdot,\cdot]$ represents the concatenation operation, and $\theta(\cdot)$ represents a modeling function to inference the attention vector $\boldsymbol{a}^{i} \in \mathbb{R}^{C}$ corresponding to $\boldsymbol{x}^{i}$. We define $\phi(\boldsymbol{x}^{i})=ReLU(W_\phi\boldsymbol{x}^{i})$, $\psi(\boldsymbol{r}^{i})=ReLU(W_\psi\boldsymbol{r}^{i})$, and $\theta([\phi(\boldsymbol{x}^{i}), \psi(\boldsymbol{r}^{i})])=ReLU(W_\theta([\phi(\boldsymbol{x}^{i}), \psi(\boldsymbol{r}^{i})]$), where $W_\phi \in \mathbb{R}^{(C/s) \times C}$, $W_\psi \in \mathbb{R}^{(D/s) \times D}$ and $W_\theta \in \mathbb{R}^{C \times ((C/s)+(D/s))}$ are learned weight matrices. We implement them by performing $1\times 1$ convolutional filtering followed by BN and ReLU. For each feature node $\boldsymbol{x}^{i}$ in $F_{all}$ (nodes corresponding to all the spatial and temporal positions), we obtain an attention score vector $\boldsymbol{a}^{i}$. For all nodes in $F_{all}$, we have $\mathcal{A} = [\boldsymbol{a}^{1},\boldsymbol{a}^{2},\cdots,\boldsymbol{a}^{K}]$.

We normalize the learned attention scores via the Softmax function across different spatial and temporal positions (node indexes) and obtain the final attention $\hat{\boldsymbol{a}}^{i}$, $i=1,2,\cdots,K$. Afterwards, we use the final attention as the weights to aggregate all the feature nodes (from all spatial and temporal positions) in $F_{all}$. Mathematically, we obtain the final sequence-level feature representation $\boldsymbol{v} \in \mathbb{R}^{C}$ by
\begin{equation}\label{eq:3}
\boldsymbol{v}=\sum_{k=1}^K\hat{\boldsymbol{a}}^{i}\odot \boldsymbol{x}^{i}, \quad  \hat{\boldsymbol{a}}^{i}=Softmax(\boldsymbol{a}^{i}),
\end{equation}
where symbol $\odot$ represents element-wise multiplication.

\subsection{Multi-Granularity Attention}
\label{subsec:MG}

Human can capture the different semantics (such as gender, body shape, clothing details) of a person at different granularity levels (\eg, in terms of viewing distance or image resolution). Some types of semantics (\eg, whether the person wears glasses or not) may be easier to capture at fine granularity while some others (\eg, body shape) may be easier to  capture at coarse granularity by excluding the distraction from fine details. Motivated by this, we propose the Multi-Granularity Reference-aided Attentive Feature Aggregation (MG-RAFA) which derives the attention and introduces a hierarchical design, aiming at capturing the discriminative spatial and temporal information at different semantic levels. Basically, we distinguish the different granularities by modeling relations and deriving attention on feature maps of different resolutions. 

Following the earlier notations, for both reference nodes in $F_R$ and the nodes to be aggregated in $F_{all}$, we split them along their channel dimensions into $N$ splits/groups. Each group corresponds to a granularity. In this way, we reduce the computational complexity in comparison with the single granularity case. For the $m^{th}$ granularity, we perform spatial average pooling with a ratio factor $m$ on both the $m^{th}$ split features of $F_R$ and $F_t$, $t=1, 2, \cdots, T$. We obtain the factorized reference feature $F_{R,m} \in \mathbb{R}^{{H_m} \times {W_m} \times \frac{C}{N}}$ of $D_m = {H_m} \times {W_m}$ nodes, where $H_m = \frac{H}{2^{m-1}}$ and $W_m = \frac{W}{2^{m-1}}$. Similarly, we obtain the factorized feature map on frame $t$ as $F_{t,m} \in \mathbb{R}^{{H_m} \times {W_m} \times \frac{C}{N}}$ and the spatial and temporal feature node set as $F_{all,m} = \{F_{t,m}| t=1, 2, \cdots, T\}$. 

Then, we employ the reference-aided attentive feature aggregation as described in Section \ref{subsec:rga} for each group separately. Thereby, the relation modeling in Eq. (\ref{eq:relation}) and the attention modeling function in Eq. (\ref{eq:attentionscore}) can be expanded into their multi-granularity versions as 
\begin{equation}\label{eq:4}
\vspace{-3mm}
r(\boldsymbol{x}_{m}^{i}, \boldsymbol{y}_{m}^{j})=\mu_m(\boldsymbol{x}_{m}^{i})^{T}\cdot\nu_m(\boldsymbol{y}_{m}^{j}),
\end{equation}
\begin{equation}\label{eq:5}
\boldsymbol{a}_{m}^{i}=\theta_m([\phi_m(\boldsymbol{x}_{m}^{i}), \psi_m(\boldsymbol{r}_{m}^{i})]),
\end{equation}
where subscript $m$ identifies the index of granularity, $\boldsymbol{x}_{m}^{i}$ denotes the $i^{th}$ node in $F_{all,m}$ and $\boldsymbol{y}_{m}^{j}$ denotes the $j^{th}$ nodes in the reference feature  map $F_{R,m}$. Similar to the feature aggregation under single granularity in Section \ref{subsec:rga}, we normalize the attention scores via Softmax function and weighted sum the feature nodes (across different spatial-temporal positions). Finally, we concatenate the fused feature of each split/group (denoted by $\boldsymbol{v}_m$) to obtain the final sequence-level feature representation $\boldsymbol{v} = [\boldsymbol{v}_1, \boldsymbol{v}_2, \cdots, \boldsymbol{v}_N]$.

\subsection{Loss Design}
\label{loss}

We add the retrieval-based loss, \ie, the triplet loss with hard mining $L_{Tr}$, and the ID/classification loss (cross entropy loss with label smoothing \cite{szegedy2016rethinking}) denoted by $L_{ID}$, on the video feature vector $\boldsymbol{v}$. Each classifier consists of a Batch Normalization (BN) layer followed by a fully-connected (FC) layer. Specially, to encourage the network to aggregate discriminative features at each granularity, we add the two loss functions on the aggregated feature of each granularity $\boldsymbol{v_g}, g=1,\cdots,N$. The final loss is:
\begin{equation}\label{eq:7}
L \!=\! L_{ID}(\boldsymbol{v}) \!+\! L_{Tr}(\boldsymbol{v}) \!+\! \frac{1}{N}\sum_{g=1}^{N} \left( L_{ID}(\boldsymbol{v}_g) \!+\! L_{Tr}(\boldsymbol{v}_g) \right).
\end{equation}

\section{Experiments}

\subsection{Datasets and Evaluation Metrics}

\setlength{\tabcolsep}{6.5 pt}
\begin{table}[!htb]
  \centering
  \caption{Three public datasets for video-based person reID.}
  \footnotesize
    \begin{tabular}{rccc}
    \hline
    \multicolumn{1}{l}{Datasets} & \multicolumn{1}{|c}{MARS \cite{zheng2016mars}}  & \multicolumn{1}{|c}{iLIDS-VID \cite{wang2014person}} & \multicolumn{1}{|c}{PRID2011 \cite{hirzer2011person}} \bigstrut[tb]\\
    \hline
    \multicolumn{1}{l}{Identities} & \multicolumn{1}{|c}{1261}  & \multicolumn{1}{|c}{300} & \multicolumn{1}{|c}{200} \bigstrut[tb]\\
    \hline
    \multicolumn{1}{l}{Tracklets} & \multicolumn{1}{|c}{20751} & \multicolumn{1}{|c}{600} & \multicolumn{1}{|c}{400} \bigstrut[tb]\\
    \hline
    \multicolumn{1}{l}{Distractors} & \multicolumn{1}{|c}{3248 tracklets} & \multicolumn{1}{|c}{0} & \multicolumn{1}{|c}{0} \bigstrut[tb]\\
    \hline
    \multicolumn{1}{l}{Cameras} & \multicolumn{1}{|c}{6}  & \multicolumn{1}{|c}{2} & \multicolumn{1}{|c}{2} \bigstrut[tb]\\
    \hline
    \multicolumn{1}{l}{Resolution} & \multicolumn{1}{|c}{$128\times 256$} & \multicolumn{1}{|c}{$64\times 128$} & \multicolumn{1}{|c}{$64\times 128$} \bigstrut[tb]\\
    \hline
    \multicolumn{1}{l}{Box Type} & \multicolumn{1}{|c}{detected} & \multicolumn{1}{|c}{manual} & \multicolumn{1}{|c}{manual} \bigstrut[tb]\\
    \hline
    \multicolumn{1}{l}{Evaluation} & \multicolumn{1}{|c}{CMC \& mAP} & \multicolumn{1}{|c}{CMC} & \multicolumn{1}{|c}{CMC} \bigstrut[tb]\\
    \hline
    \end{tabular}
  \label{tab:datasets}%
  \vspace{-2mm}
\end{table}%
\setlength{\tabcolsep}{6.6pt}
\begin{table*}[t]
  \centering
  \caption{The ablation study for our proposed multi-granularity reference-aided global attention (MG-RAFA) module. Here,``SG'' denotes ``Single-Granularity'' and ``MG'' denotes "Multi-Granularity". $N$ denotes the number of granularities. $S$ denotes the number of splits(groups) along the channel dimension for masking attention on each split respectively. In a multi-granularity setting, the number of splits is equal to the number of granularities (\ie, $S=N$) since each split correponds to a granularity level. We use ``MG-AFA'' to represent the attention module without relations, in which attention values are inferred from RGB information alone.}
  \footnotesize
    \begin{tabular}{r|c|cccc|ccc|ccc}
    \toprule
    \multicolumn{1}{c|}{\multirow{2}[2]{*}{Models}} & \multicolumn{1}{c|}{\multirow{2}[2]{*}{\#GFLOPs}} & \multicolumn{4}{c|}{Mars}      & \multicolumn{3}{c|}{iLIDS-VID} & \multicolumn{3}{c}{PRID2011} \\
    \multicolumn{1}{c|}{} & \multicolumn{1}{c|}{} & \,\,mAP\,\,   & Rank-1   & Rank-5   & Rank-20  & Rank-1   & Rank-5   & Rank-20  & Rank-1   & Rank-5   & Rank-20 \\
    \midrule
    \multicolumn{1}{l|}{Baseline} & 32.694 & 82.1  & 85.9  & 95.1  & 97.3  & 86.5  & 96.6  & 98.9  & 92.5  & 98.5  & 99.6  \\
    \multicolumn{1}{l|}{MG-AFA (\emph{N=4})} & +0.095 & 82.5  & 86.6  & 96.1  & 97.8  & 86.7  & 96.6  & 98.7  & 92.6  & 98.1  & 99.6 \\
    \hline
    \multicolumn{1}{l|}{SG-RAFA (\emph{S=1})} & +2.301 & 85.1  & 87.8  & 96.1  & \textbf{98.6}  &  87.1 & 97.1  & 99.0  & 93.6  & 98.2  & 99.9 \\
    \multicolumn{1}{l|}{SG-RAFA (\emph{S=4})} & +0.615 & 84.9  & 88.4  & 96.6  & 98.5  & 86.7  & 96.6  & 98.7  & 94.2  & 98.6  & 99.6 \\
    \hline
    \multicolumn{1}{l|}{MG-RAFA (\emph{N=2})} & +0.742 & 85.5  & 88.4  & \textbf{97.1}  & 98.5  & 87.1  & 97.3  & 99.3  & 94.2  & 98.2  & 99.9  \\
    \multicolumn{1}{l|}{MG-RAFA (\emph{N=4})} & +\textbf{0.212} & \textbf{85.9}  & \textbf{88.8}  & 97.0  & 98.5  & \textbf{88.6}  & \textbf{98.0}  & \textbf{99.7}  & \textbf{95.9}  & \textbf{99.7}  & \textbf{100} \\
    \bottomrule
    \end{tabular}
  \label{tab:ablation}%
\end{table*}%
We evaluate our approach on three video-based person reID datasets, including MARS \cite{zheng2016mars}, iLIDS-VID \cite{wang2014person}, and PRID2011 \cite{hirzer2011person}. Table \ref{tab:datasets} gives detailed information. Following the common practices, we adopt the Cumulative Matching Characteristic (CMC) at Rank-1 (R-1), to Rank-20 (R-20), and the mean average precision (mAP) as the evaluation metrics. For MARS, we use the  train/test split protocol defined in ~\cite{zheng2016mars}. For iLIDS-VID and PRID2011, similar to \cite{liu2018video, chen2018video, li2018diversity}, we report the average CMC across 10 random half-half train/test splits for stable comparison. 

\vspace{-1mm}
\subsection{Implementation Details}

\noindent\textbf{Networks.} Similar to \cite{almazan2018re,luo2019bags,fu2019sta}, we take ResNet-50 \cite{he2016deep} as our backbone for per-frame feature extraction. Similar to \cite{sun2017beyond,zhang2019densely}, we remove the last spatial down-sampling operation in the conv5\_x block for both the baseline and our schemes. In our scheme, we apply our propose MG-RAFA after the last residual block (conv5\_x) for the attentive feature aggregation to obtain the final feature vector $\boldsymbol{v}$. We build \emph{Baseline} by taking the feature vector $\boldsymbol{v}$ obtained through global spatial-temporal average pooling.

\noindent\textbf{Experimental Settings.} We uniformly split the entire video into \emph{T=8} chunks and randomly sample one frame per chunk. For the triplet loss with hard mining \cite{hermans2017defense}, in a mini-batch, we sample $P$ identities and each identity includes $Z$ different video tracklets. For MARS, we set $P$=16 and $Z$=4 so that the mini-batch size is 64. For iLIDS-VID and PRID2011, we set $P$=8 and $Z$=4. We use the commonly used data augmentation strategies of random cropping \cite{wang2018resource}, horizontal flipping, and random erasing \cite{zhong2017random,wang2018resource,wang2018mancs} (with a probability of 0.5 \emph{at the sequence level} for both the baselines and our schemes. Sequence-level data augmentation is much superior to frame-level one. This is closer to the realistic data variation and
will not break the inherent consistency among frames. We set the input  resolution of images to be $256 \times 128$ with $T$=8 frames. Adam optimizer is used. Please find more details in the supplementary.


\subsection{Ablation Study}

\vspace{-1mm}
\subsubsection{Effectiveness Analysis} 
\vspace{-2mm}

We validate the effectiveness of our proposed multi-granularity reference-aided attention (MG-RAFA) module and show the comparisons in Table \ref{tab:ablation}.

\noindent\textbf{MG-RAFA vs. Baseline.} Our final scheme \emph{MG-RAFA ($N\!=\!4$)} outperforms \emph{Baseline} by \textbf{2.9\%}, \textbf{2.1\%} and \textbf{3.4\%} on Mars, iLIDS-VID, and PRID2011, respectively. This demonstrates the effectiveness of our proposed attentive aggregation approach. 


\noindent\textbf{Effectiveness of using the Global Relations.} In Table \ref{tab:ablation}, \emph{MG-AFA~(N=4)} denotes the scheme when we use the visual feature alone (without using relations) to learn the attention. Our scheme \emph{MG-RAFA~(N=4)} which uses relations outperforms \emph{MG-AFA (N=4)} by 2.2\%,  1.9\%, and 3.3\% in Rank-1 on Mars, iLIDS-VID, and PRID2011, respectively, indicating the effectiveness of leveraging global relations for learning attention.

\noindent\textbf{Single Granularity vs. Multiple Granularity.} 
\emph{SG-RAFA(S=1)} also takes advantage of relations but ignores the exploration of semantics at \emph{different granularities}. In comparison with \emph{SG-RAFA(S=1)}, our final scheme \emph{MG-RAFA(N=4)} that explores relations at multiple granularities achieves 1.0\%, 1.5\%, 2.3\% Rank-1 improvements on Mars, iLIDS-VID, PRID2011, respectively. \emph{MG-RAFA} is effective in capturing correlations of different granularity levels.


Moreover, we study the effects of different numbers of granularities by comparing \emph{MG-RAFA(N=4)} with \emph{MG-RAFA(N=2)}. The results show that finer granularity delivers better performance. Note that the spatial resolution of the frame features $F_t$ is $16\times8$. The spatial resolution ratio between two adjacent granularity levels is set to 4, which allows the maximum number of granularity levels $N$ to be 4 (\ie, $16\times8$, $8\times4$, $4\times2$, and $2\times1$) in this work. In the subsequent description, we use \emph{MG-RAFA} to refer to \emph{MG-RAFA(N=4)} unless otherwise specified.

To further demonstrate that the improvements come from the relation modeling at varying granularities instead of multiple attention masks, we compare \emph{MG-RAFA(N=4)} with the single-granularity setting \emph{SG-RAFA(S=4)}. In this setting, the features are divided into four splits(groups) along channels with each split having an attention mask rather than a shared one. Each attention mask is derived from the same fine gruanularity. The results show that \emph{MG-RAFA(N=4)} is superior to \emph{SG-RAFA(S=4)}.

\noindent\textbf{Complexity.} Thanks to the channel splitting and spatial pooling, the computational complexity (FLOPS) of our multi-granularity  module \emph{MG-RAFA(N=4)} is only 9.2\% of that of the single granularity module \emph{SG-RAFA(S=1)}.

\vspace{-4mm}
\subsubsection{Selection of the Reference Feature Nodes}
\vspace{-2mm}

In our scheme, we take a set of feature nodes (S-RFNs) as reference to model pairwise relations. Instead of taking all feature nodes in the frame features as reference, considering the larger temporal redundancy, an average pooling operation along the time dimension is performed to reduce the number of nodes for easing optimization and reducing computational complexity. For clarity, we investigate different strategies for building the S-RFNs under single granularity setting (\ie, SG-RAFA) and show the results in Table \ref{tab:global-feature}.

\setlength{\tabcolsep}{3pt}
\begin{table}[t]
  \centering
  \caption{Comparison of different strategies on selection of the reference feature nodes (S-RFNs). Different spatial(S) and temporal (T) pooling strategies are compared. We denote the spatial and temporal node dimension as $(H\times W\times T)$. For example, we build the S-RFNs by adopting temporal average pooling, leading to $16\times 8\times 1$ nodes in S-RFNs.}
  
  \footnotesize
    \begin{tabular}{rccccccc}
    \toprule
    \multicolumn{1}{c}{\multirow{2}[4]{*}{S-RFNs}} & \multicolumn{1}{c}{\multirow{2}[4]{*}{\#Nodes}} & \multicolumn{1}{c}{\multirow{2}[4]{*}{\#GFLOPs}} & \multicolumn{5}{c}{Mars} \\
    \cmidrule{4-8}
    \multicolumn{1}{c}{} & \multicolumn{1}{c}{} & \multicolumn{1}{c}{} & mAP  & R-1   & R-5   & R-10  & R-20 \\
    \midrule
    \multicolumn{1}{l}{Baseline} & 0 & 32.694 & 82.1  & 85.9  & 95.1  & 96.5  & 97.3 \\
    \midrule
    \multicolumn{1}{l}{S-P (8$\times$1$\times$8)} & 64 & +2.034 & 83.9  & 86.6  & 96.1  & 97.4  & 98.0 \\
    \multicolumn{1}{l}{S-P (8$\times$2$\times$8)} & 128 & +2.345 & 83.9  & 87.2  & 95.6  & 97.2  & 97.9 \\
    \multicolumn{1}{l}{S-P (4$\times$4$\times$8)} & 128 & +2.345 & 84.1  & 87.1  & 95.7  & 97.3  & 97.9 \\
    \multicolumn{1}{l}{S-P (8$\times$4$\times$8)} & 256 & +2.967 & 84.2  & 87.0  & 95.8  & 97.4  & 98.0 \\
    \multicolumn{1}{l}{T-P (16$\times$8$\times$2)} & 256 & +2.916 & 84.7  & 87.4  & 96.1  & 97.4  & 98.3 \\
    \multicolumn{1}{l}{T-P (16$\times$8$\times$4)} & 512 & +4.159 & 84.7  & 87.3  & 96.1  & 97.4  & 98.1 \\
    \multicolumn{1}{l}{ST (16$\times$8$\times$8)} & 1024 & +6.697 & 84.3  & 87.3  & 95.8  & 97.2  & 98.1 \\
    \midrule
    \multicolumn{1}{l}{Ours (16$\times$8$\times$1)} & 128 & +2.301 & \textbf{85.1}  & \textbf{87.8}  & \textbf{96.1}  & \textbf{97.8}  & \textbf{98.5} \\
    \bottomrule
    \end{tabular}
  \label{tab:global-feature}%
  \vspace{-2mm}
\end{table}%

The spatial resolution of frame feature is $H\times W$, where \emph{H=16, W=8} in this paper, resulting in $128$ feature nodes for each frame feature $F_t$. The number of temporal  frames is \emph{T=8}. 
\textbf{\emph{Ours}}: we obtain S-RFNs by fusing frame features $F_t$, $t=1,\cdots,T$ along the time dimension and obtain a feature map with $16 \times 8 = 128$ feature nodes. \textbf{\emph{S-P}}: we fuse feature nodes to obtain the reference set via average pooling along spatial dimensions. \textbf{\emph{T-P}}: we perform average pooling along the temporal dimension with different ratios to obtain different settings in Table \ref{tab:global-feature}. \textbf{\emph{ST($16\times8\times8$)}}: we take all the spatial and temporal nodes as the reference set.

We have the following observations. (1) \emph{Ours} outperforms schemes \emph{S-P($8\times1\times8$)}, \emph{S-P($8\times2\times8$)}, S-P($4\times4\times8$), \emph{S-P($8\times4\times8$)} with spatial pooling by 1.2\%, 1.2\%, 1.0\% and 0.9\% in mAP respectively, where spatial pooling may remove too much useful information and result in inferior S-RFNs. (2) \emph{Ours} also outperforms those with partial temporal pooling. The performance increases as the temporal pooling degree increases. (3) \emph{Ours} outperforms the scheme \emph{ST($16\times8\times8$)} without any pooling by 0.8\% in mAP. Using all nodes as reference results in a larger optimization space and the diversity of temporal patterns is complex which makes it difficult to learn. In contrast, through temporal average pooling, we reduce the pattern complexity and thus ease the learning difficulty and computational complexity.

\noindent\textbf{Complexity.} Thanks to the selection of S-RFNs, in comparison with \emph{ST($16\times8\times8$)} which uses all the feature nodes as reference, the computational complexity in terms of FLOPs for our aggregation module is reduced from 6.697G to 2.301G while the performance is 0.8\% higher in mAP.

\setlength{\tabcolsep}{10pt}
\begin{table}[t]
  \centering
  \caption{Comparison with non-local related schemes.}
  \footnotesize
    \begin{tabular}{rccccc}
    \toprule
    \multicolumn{1}{c}{\multirow{2}[4]{*}{Models}} & \multicolumn{5}{c}{Mars} \\
    \cmidrule{2-6}
    \multicolumn{1}{c}{} & mAP   & R-1   & R-5   & R-10  & R-20 \\
    \midrule
    \multicolumn{1}{l}{NL(S)} & 83.2  & 86.6  & 95.9  & 97.1  & 97.9 \\
    \multicolumn{1}{l}{NL(ST)} & 82.7  & 86.0  & 95.4  & 96.7  & 97.4 \\
    \midrule
    \multicolumn{1}{l}{SG-RAFA} & \textbf{85.1}  & \textbf{87.8}  & \textbf{96.1}  & \textbf{97.8}  & \textbf{98.5} \\
    \bottomrule
    \end{tabular}
  \label{tab:non-local}
  \vspace{-2mm}
\end{table}

\setlength{\tabcolsep}{8pt}
\begin{table}[t]
  \centering
  \caption{Evaluation of the multi-granularity (MG) design when other attention methods are used on the extracted feature maps $F_t, t=1,\cdots, T$. Granularity is set to $N=4$.}
  \footnotesize
    \begin{tabular}{rccccc}
    \toprule
    \multicolumn{1}{c}{\multirow{2}[4]{*}{Models}} & \multicolumn{5}{c}{Mars} \\
    \cmidrule{2-6}
    \multicolumn{1}{c}{} & mAP   & R-1   & R-5   & R-10  & R-20 \\
    \midrule
    \multicolumn{1}{l}{Baseline} & 82.1  & 85.9  & 95.1  & 96.5  & 97.3 \\
    \midrule
    \multicolumn{1}{l}{RGA-SC} & 83.5  & 87.2  & 95.3  & 97.1  & 98.2 \\
    \multicolumn{1}{l}{RGA-SC (MG)} & \textbf{85.0}    & \textbf{88.1}  & \textbf{96.9}  & \textbf{97.7}  & \textbf{98.5} \\
    \midrule
    \multicolumn{1}{l}{SE}   & 82.9  & 86.5  & 95.5  & 97.1  & 98.1 \\
    \multicolumn{1}{l}{SE (MG)} & \textbf{84.3}  & \textbf{87.6}  & 95.4  & 97.1  & 98.1 \\
    \midrule
    \multicolumn{1}{l}{CBAM}  & 82.9  & 86.8  & 95.7  & 97.2  & 98.1 \\
    \multicolumn{1}{l}{CBAM (MG)} & \textbf{84.6}  & \textbf{88.0} & 95.7  & 97.2  & 97.9 \\
    \midrule
    \multicolumn{1}{l}{MG-RAFA (Ours)} & \textbf{85.9}  & \textbf{88.8}  & \textbf{97.0} & \textbf{97.7}  & \textbf{98.5} \\
    \bottomrule
    \end{tabular}
  \label{tab:other-attention}%
  \vspace{-3mm}
\end{table}%

\setlength{\tabcolsep}{7pt}
\begin{table*}[t]
  \centering
  \caption{Performance (\%) comparison of our scheme with the state-of-the-art methods on three benchmark datasets\protect\footnotemark[2].}
  \footnotesize
    \begin{tabular}{rcccc|ccc|ccc}
    \toprule
    \multicolumn{1}{c|}{\multirow{2}[2]{*}{Models}} & \multicolumn{4}{c|}{Mars} & \multicolumn{3}{c|}{iLIDS-VID} & \multicolumn{3}{c}{PRID2011} \\
    \multicolumn{1}{c|}{} & \,\,\,mAP\,\,   & Rank-1   & Rank-5   & Rank-20  & Rank-1   & Rank-5   & Rank-20  & Rank-1   & Rank-5   & Rank-20 \\
    \midrule
    \multicolumn{1}{l|}{AMOC (TCSVT17)\cite{liu2018video}} & 52.9  & 68.3  & 81.4  & 90.6  & 68.7  & 94.3  & 99.3  & 83.7  & 98.3  & \textbf{100} \\
    \multicolumn{1}{l|}{TriNet (ArXiv17)\cite{hermans2017defense}} & 67.7  & 79.8  & 91.4  & - & - & - & - & - & -  & - \\
    \multicolumn{1}{l|}{3D-Conv+NL (ACCV18)\cite{liao2018video}} & 77.0  & 84.3  & -  & -  & 81.3 & - & - & 91.2  & -  & -\\
    \multicolumn{1}{l|}{Snippt (CVPR18)\cite{chen2018video}} & 76.1  & 86.3  & 94.7  & 98.2  & 85.4  & 96.7  & \underline{99.5}  & 93.0 & 99.3 & \textbf{100} \\
    \multicolumn{1}{l|}{DRSA (CVPR18)\cite{li2018diversity}} & 65.8  & 82.3  & - & - & 80.2  & - & - & 93.2  & - & - \\
    \multicolumn{1}{l|}{DuATM (CVPR18)\cite{si2018dual}} & 62.3  & 78.7  & - & - & - & - & - & - & - & - \\
    \multicolumn{1}{l|}{M3D (AAAI19)\cite{li2019multi}} & 74.1  & 84.4  & 93.8  & 97.7  & 74.0 & 94.3 & - & 94.4 & \textbf{100} & - \\
    \multicolumn{1}{l|}{STA (AAAI19)\cite{fu2019sta}} & \underline{80.8}  & 86.3  & 95.7  &  -  & - & - & - & - & - &  \\
    \multicolumn{1}{l|}{Attribute (CVPR19)\cite{zhao2019attribute}} & 78.2  & 87.0  & 95.4  &  \textbf{98.7}  &  \underline{86.3} & \underline{97.4} & \textbf{99.7} & 93.9 & 99.5 & \textbf{100} \\
    \multicolumn{1}{l|}{GLTR (ICCV19)\cite{li2019global}} & 78.5  & \underline{87.0}  & \underline{95.8}  &  98.2  &  86.0 & \textbf{98.0} & - & \underline{95.5} & \textbf{100} & - \\
    \midrule
    \multicolumn{1}{l|}{MG-RAFA (Ours)} & \textbf{85.9}  & \textbf{88.8}  & \textbf{97.0}  & \underline{98.5}  & \textbf{88.6}  & \textbf{98.0}  & \textbf{99.7}  & \textbf{95.9}  & \underline{99.7}  & \textbf{100} \\
    \bottomrule
  \end{tabular}
  \label{tab:sota}%
  \vspace{-2mm}
\end{table*}%

\vspace{-4mm}
\subsubsection{Comparison with Non-local}
\vspace{-2mm}

Non-local block \cite{wang2018non} explores long-range context, which weighted sums the features from all positions to refine the current position feature. Both our approach and non-local can explore the global context. However, non-local block uses a \emph{deterministic} way, \ie, weighted summation (without parameters), to exploit the global information, which limits its capability. In contrast, ours could mine the structural pattern and semantics from the stacked relations by leveraging a learned model/function to inference the importance of a feature node as attention, being more flexible and having large optimization space. 

Table \ref{tab:non-local} shows the comparisons to non-local schemes. We added a non-local module on the feature maps $F_t, t=1,\cdots,T$ for feature refinement followed by spatio-temporal average pooling. \emph{NL(ST)} denotes that non-local is performed on all the spatio-temporal features and \emph{NL(S)} denotes that non-local is performed within each frame. Our \emph{SG-RAFA} outperforms \emph{NL(ST)} and \emph{NL(S)} significantly by 2.4\% and 1.9\% in mAP, respectively. \emph{NL(ST)} is inferior to \emph{NL(S)} which may be caused by the optimization difficulty when spatio-temporal dimensions are jointly considered.

\vspace{-5mm}
\subsubsection{Extension of MG Design to Other Attention}
\vspace{-2mm}

Different semantics could be suitably captured at different granularities (as illustrated in Figure \ref{fig:motivation}). Our proposed multi-granularity design can also be applied to other attention mechanisms. We conduct experiments by applying several different attention designs on the extracted feature maps $F_t, t=1,\cdots, T$ and show the results in Table \ref{tab:other-attention}. Compared with the single gruanularity versions, multi-granularity design brings 1.5\%, 1.4\%, and 1.7\% improvements in mAP respectively for RGA-SC \cite{zhang2019relation}, SE \cite{hu2018squeeze}, and CBAM \cite{woo2018cbam}, demonstrating the effectiveness of the proposed multi-granularity design. In addition, our proposed MG-RAFA outperforms  \emph{RGA-SC(MG)}, \emph{SE(MG)}, \emph{CBAM(MG)} by 0.9\%, 1.6\%, and 1.3\% in mAP respectively.

\vspace{-2mm}
\subsection{Comparison with State-of-the-arts}

Table \ref{tab:sota} shows that our \emph{MG-RAFA} significantly outperforms the state-of-the-arts. On Mars, compared to STA \cite{fu2019sta}, our method achieves \textbf{5.1\%} improvements in mAP. On iLIDS-VID and PRID2011, ours outperforms the second best approach by 2.3\% and 0.4\% in Rank-1, respectively. 


\vspace{-2mm}
\subsection{Visualization Analysis}

\begin{figure}[t]
\vspace{-2mm}
\centering
\vspace{-1mm}
\subfigure[Visualization on different frames at the $2^{nd}$ granularity.] 
{\includegraphics[width=1\linewidth]{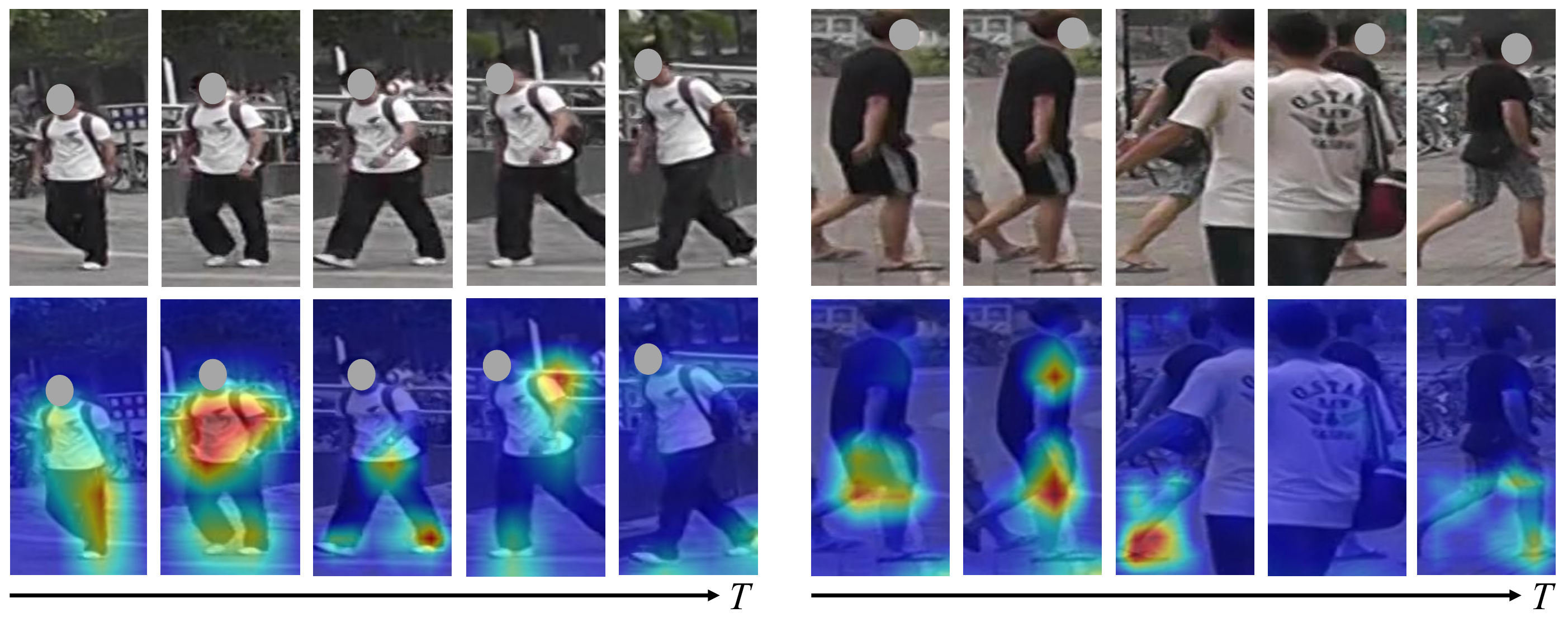}}
\subfigure[Visualization of different granularities at a given time.]
{\includegraphics[width=1\linewidth]{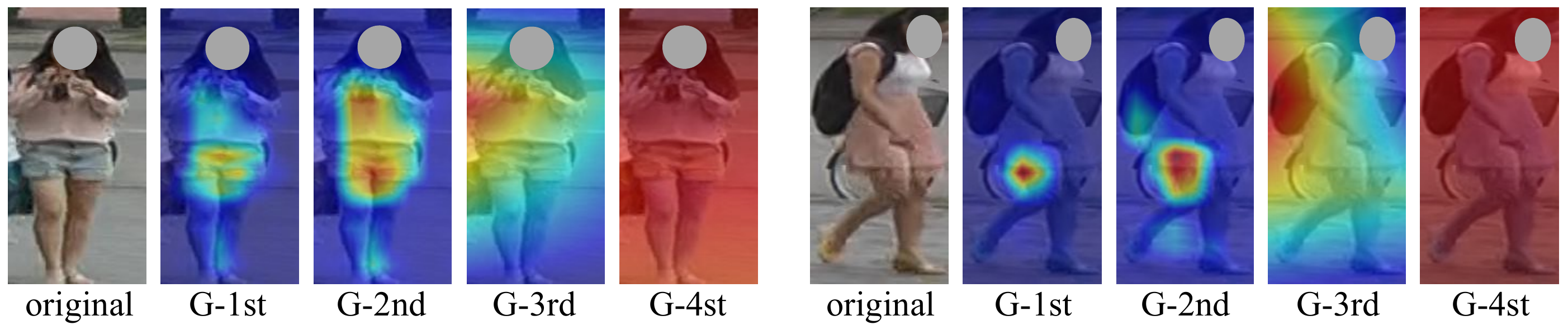}}
\caption{Visualization of our attention (a) across different frames, and (b) at different granularities. ``G-1st'' to ``G-4th'' denote the $1st$ to the $4th$ granularities, with their corresponding spatial resolutions of the attention masks for each frame as 16$\times$8, 8$\times$4, 4$\times$2, 2$\times$1, respectively. Here, we rescale the attention map of different spatial resolutions to the same spatial resolution for visualization.}
\label{fig:visualization}
\vspace{-4mm}
\end{figure}

We visualize the learned attention values at different spatial-temporal positions at different granularities in Figure \ref{fig:visualization}. We have two observations from (a). (1) The learned attention tends to focus on different semantic regions from different frames, which gets rid of a lot of repetitions (redundancy). (2) Interestingly but not surprisingly, our attention is able to select the better represented areas and exclude the inferences (\eg, see the $3^{rd}$ and $4^{th}$ columns of the right sub-figures in (a) where there are occlusions). We believe our attention modeling is an effective method to capture and learn discriminative spatial and temporal representation. (b) shows MG-RAFA captures different semantics at different granularities, which tends to capture more details at finer granularities and larger body parts at coarser granularities.


\footnotetext[2]{We do not include results on DukeMTMC-VideoReID \cite{zheng2017unlabeled} since this dataset is not publicly released anymore.}

\vspace{-3mm}
\section{Conclusion}
In this paper, we propose a Multi-Granularity Reference-aided Attentive Feature Aggregation scheme (MG-RAFA) for video-based person re-identification, which effectively enhances discriminative features and suppresses identity-irrelavant features on the spatial and temporal feature representations. Particularly, to reduce the optimization difficulty, we propose to use a representative set of reference feature nodes (S-RFNs) for modeling the global relations. Moreover, we propose multi-gruanularity attention by exploring the relations at different granularity levels to capture semantics at different levels. Our scheme achieves the state-of-the-art performance on three benchmark datasets. 

\section*{Acknowledgements}
This work was supported in part by NSFC under Grant U1908209, 61632001 and the National Key Research and Development Program of China 2018AAA0101400.

{\small
\bibliographystyle{ieee_fullname}
\bibliography{main}
}

\end{document}